# AUTONOMOUS REQUIREMENTS SPECIFICATION PROCESSING USING NATURAL LANGUAGE PROCESSING

Professor S.G. MacDonell
Software Engineering Research Lab
Auckland University of Technology
Auckland, New Zealand
stephen.macdonell@aut.ac.nz

Dr Kyongho Min
School of Computer & Info Sciences
Auckland University of Technology
Auckland, New Zealand
kyongho.min@aut.ac.nz

Dr A.M. Connor
Software Engineering Research Lab
Auckland University of Technology
Auckland, New Zealand
andrew.connor@aut.ac.nz

**Abstract**

We describe our ongoing research that centres on the application of natural language processing (NLP) to software engineering and systems development activities. In particular, this paper addresses the use of NLP in the requirements analysis and systems design processes. We have developed a prototype toolset that can assist the systems analyst or software engineer to select and verify terms relevant to a project. In this paper we describe the processes employed by the system to extract and classify objects of interest from requirements documents. These processes are illustrated using a small example.

## 1 INTRODUCTION

This paper describes the architecture of an autonomous requirements specification processing system that utilises a limited version of a natural language processing (NLP) system and an interactive user interface system. When analyzing requirements artefacts e.g. specification documents, interview transcripts and so on, an analyst generally uses their own software engineering knowledge, training and experience in combination with one or more software design tools. In particular, however, the *verification* of requirements specification analysis depends primarily on the software engineer's knowledge. As a result, important information such as relationships between entities in a requirements specification document could possibly be missed.

It is rather stating the obvious, but the requirements analysis and determination activities are among the most important in information systems development. Inaccuracies that are introduced or omissions that occur in these stages of development, if unchecked, generally result in costly rework in later lifecycle phases. The work described in this paper is therefore focused on the verification of requirements specification analysis as performed by a software engineer or systems analyst with a view to producing a design model – a use case diagram, an entity-relationship model or similar. This paper first describes prior autonomous application research in requirements analysis in section 2. This is followed by a description of the proposed system architecture in section 3. Section 4 closes the paper with a brief discussion and our conclusions to date.

## 2 BACKGROUND

Many of the problems encountered in software systems can be traced back to shortcomings in the processes and practices used to gather, specify and manage the end product requirements. Typically, these shortcomings are due to the use of informal information gathering, unstated or implicit functionality, unfounded or uncommunicated assumptions, inadequately documented requirements or a casual requirements change process [1]. It has been suggested that between 40 and 60% of software defects are related to errors made during the requirements stage [2]. The cost of correcting defects is often significantly greater than the cost that would have been incurred to ensure that the requirements correctly represented the users' need.

Whilst the generation of a complete and non-ambiguous set of requirements reduces the risk in any given project, there is still a risk that the requirement set is not transformed into an appropriate design. This risk is inherent as a result of mis-interpretation of the requirements, particularly due to a lack of shared understanding [3] or due to poor structuring of the project by not conducting architectural design in parallel with requirements capture [4].

The use of formal languages or a structured system design approach can greatly increase the chance that the software as constructed will in fact conform to the interpretation of the requirements. Formal languages help remove some elements of ambiguity from the process as they use explicit syntax and semantics that define a set of relations and object interactions more consistently than the English language. However, the extraction of entity relationships from a natural language requirements document is normally conducted manually by a designer using their software engineering knowledge in conjunction with a design tool. This introduces the risk of inconsistency in approach and also the possibility that some entities, relationships or attributes will be missed entirely.

A great deal of research has focused on the automation of aspects of the software engineering process, namely requirements elicitation, translation and analysis, and subsequent software generation, demonstration and test, resulting in a final system artefact. To date there have been few attempts to automate the translation from a requirements document written in a natural language to one expressed in a formal specification language. One of the major reasons for this is the ambiguity of natural language requirements.

Nazlia *et al* [5] propose new heuristics that assist the semi-automated generation of entity relationship diagrams for database modelling from a natural language description, with reasonable success. However, the limitation to database systems does imply that the natural language documents being processed have particular structure and language and their approach may not be extendable to generic software requirement documents.

Bras and Toussiant [6] specify a framework for the analysis and mapping of requirements documents, with a particular focus on satellite ground support systems. Such systems tend to be large, take a long time to develop, and have extensive documentation that is all predominantly in natural language. They facilitate requirements traceability by building tools to analyze, linguistically map and retain as a knowledge base the contents of the requirements documents.

Lee and Bryant [7] developed a system for mapping natural language requirements documents into an object-oriented formal specification language that utilises Contextual Natural Language Processing (CNLP) to overcome the ambiguity in natural language. The mapping process requires that the requirements specification is converted to an XML format which is then parsed, with the results added to a knowledge base. The content of the knowledge based is converted into a Two Level Grammar format which is a formal requirements specification language [8]. Finally, a VDM++ model is produced that describes the software design.

Ambriola and Gervasi [9] describe a system for supporting natural language requirements gathering, elicitation, selection and validation. Central to the work is the idea that requirements are supplemented by a glossary describing and classifying all the domain and system specific terms used in the requirements. Therefore, the NLP engine has *a-priori* knowledge relevant to the content of the requirements documents.

The approach detailed in this paper has no *a-priori* knowledge with regards the content of the documents, which also require no pre-processing. It is applicable to all software requirements documents as it is primarily used interactively and as such provides a high level of consistency checking to ensure that all requirements are captured in terms of the relationships between entities.

## 3 SYSTEM DESIGN

In this section, the architecture of an NL (natural language)-based SE tool is described. The system focuses on the automatic extraction of objects of interest from a requirements specification document that is being processed by a systems analyst (Figure 1).

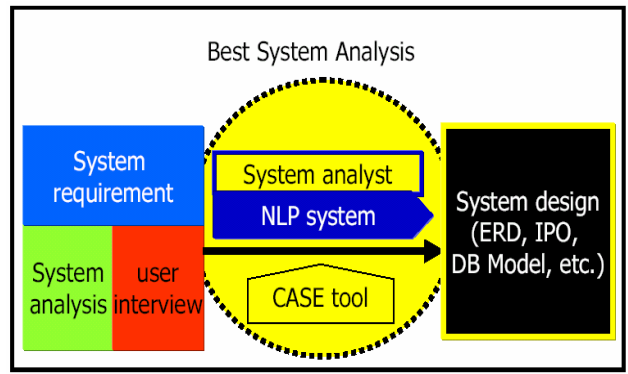

Figure 1: Assisted Requirements Analysis Process
(as implemented in this research project)

### 3.1 System Architecture

The system is composed of three modules with a user interface implemented by Common lisp IDE (Figure 2). The first of the three modules – a tokeniser – reads sentences from a document, the second module parses each sentence and extracts all unique noun terms (an NLP tool), and the third module – a term management system – performs 1) the filtering of unimportant terms, 2) the classification of the remaining terms into one of three categories (function, entity, or attribute), and 3) the insertion of objects of interest into a project knowledge base.

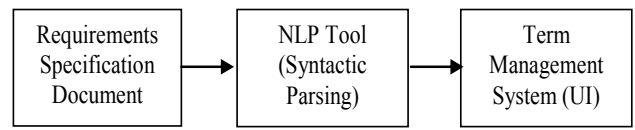

Figure 2: System Architecture

### 3.2 A Parsing System

After the sentences in a requirements specification document are extracted by the tokeniser, each sentence is parsed by a syntactic parser based on a chart parsing technique [10] with a context-free grammar (CFG) that is augmented with constraints. The current prototype system uses a dictionary with about 32000 entries and 79 rules. A context free rule is composed of LHS (Left-Hand Side), RHS (Right-Hand Side) with well-formedness constraints for the phrasal constituent. For example, there is a rule S (i.e. LHS) → NP VP (i.e. RHS) with its well-formedness constraint being (number-agreement NP VP). Thus the sentence "He see a car in the park" would be filtered out as ill-formed because of the number disagreement between "he" and "see".

At present, the syntactic parsing system does not recognise compound noun terms, such as "information system" and "staff members", by a systematical compound noun recognition system. The system recognises compound noun terms by using a list of compound noun terms and a pattern matching technique.

The syntactic parser can produce ambiguous parse trees of each sentence. At present, the parser has no disambiguation module – this will be implemented in a later version of the system. Currently the first parse tree is selected as the basis for the extraction of terms for the term management system, terms that will ultimately appear in specification and design artefacts such as use case diagrams or data models. For example, the sentence "A system requires entry of patient's information" has the following parse tree:

```
(S   (NP (DET "A") (NOUN "system"))
     (VP  (VERB "requires")
         (NP  (NP (NOUN "entry"))
             (PP (OF "of") (NP (POSSADJ "patient's")
                 (NOUN "information")))))).
```

From the parse tree, terms based on the syntactic structure (noun phrase (NP)) would be extracted. In the example above this would include (NP (DET "A") (NOUN "system")), (NP (NOUN "entry"), and (NP (POSSADJ "patient's") (NOUN "information")). However, the NP ("entry of patient's information") would not be extracted because the structure includes embedded NPs ("entry" "patient's information").

Another real, complex sentence extracted from a requirements specification document, "Dunedin Podiatry requires an information system that allows entry and retrieval of patient's details and their medical histories." results in two parse trees. From the first parse tree, the term extraction stage retrieves NOUN terms including "Dunedin Podiatry", "information system", "entry", "retrieval", "(patient's) details", and "(their medical) histories" (Figure 3).

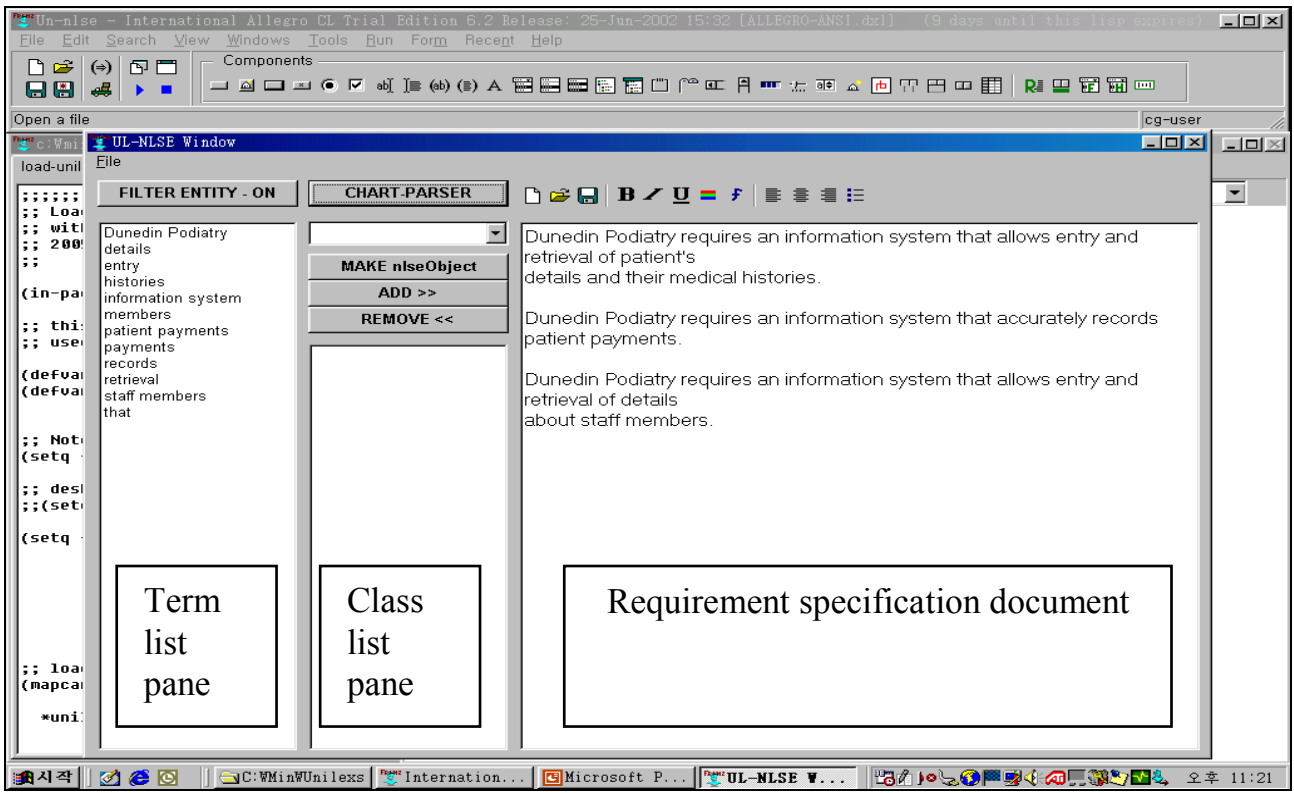

Figure 3: Term Extraction by a Syntactic Parser

Finally, the term extraction process identifies nouns in the extracted NPs, in this case nouns such as "system", "entry", and "information", and these terms can then be classified into one of the categories relevant to the design artefact being produced (e.g. entity, function, attribute) by a term management system.

### 3.3 Term Management System

After extracting NP terms, the nouns are shown in the term list pane (i.e. left pane) in Figures 3 and 4. The filtering function (enacted by the 'Filter Entity' toggle button, shown in Figure 4) enables the analyst or software engineer to remove unimportant terms. The term extraction process cannot necessarily determine every useful term automatically. Thus in this stage the user can manually remove further unimportant terms.

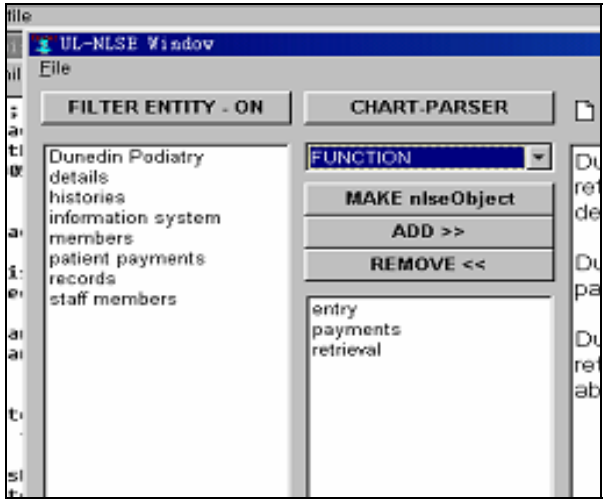

Figure 4: Filtering and categorisation of terms.

The user can then select terms to create classes of objects of interest (in this example, one of entity, attribute, or function) and can manage the term's addition to and deletion from the defined class (via the class list pane, shown as the middle pane in Figure 3 and the right-hand pane in Figure 4). The user can view the currently classified terms in each of the three classes by using a list pane of classes (i.e. a combo box under the 'chart-parser' button in Figure 4).

By selecting terms and their class, individual objects are created and stored in a project knowledge base using the following data structures:

(OBJECT (:TYPE FUNCTION) (:VALUE "entry"));
(OBJECT (:TYPE ENTITY) (:VALUE "patient")); and
(OBJECT (:TYPE ATTRIBUTE) (:VALUE "age")).

Further documents relevant to the project can then be analysed and the knowledge base updated. Class conflicts can be identified by the system and flagged to the user as requiring resolution. The knowledge base can then be used as the basis for the automatic generation of relevant design artefacts – object models, data models and the like.

## 4 DISCUSSION AND CONCLUSIONS

At present the prototype parsing system is unable to perform the following:

1. disambiguation of syntactic parse trees

2. compound noun analysis and proper noun processing

3. anaphoric resolution and semantic interpretation of terms.

The next version of the system will be extended to implement the above mentioned functionality in order to enhance the process of term extraction and enable term relationship identification. The semantic interpretation of each sentence will help in the extraction of useful relationships between the classes. For example, the parsing of "patient's medical histories" will produce in a data model a one-to-many relationship between "patient" and "medical histories".

The fully implemented system will utilise NLP to assist systems analysts in selecting and verifying objects and relationships of relevance to any given project, then enabling these objects and relationships to be depicted in design artefacts (in either this tool or additional software engineering tools). Thus the burden of analysis – requiring that the systems analyst 'parse', select and relate the objects of interest from specification documents – can be shifted at least in part to a toolset that is able to perform these tasks intelligently and automatically.